\documentclass{article}

\usepackage[accepted]{icml2023}
\usepackage{microtype}
\usepackage{graphicx}
\usepackage{subfigure}
\usepackage{booktabs} 

\usepackage{lscape}
\usepackage{array}
\usepackage{longtable}
\usepackage{multirow}
\usepackage{times}

\usepackage{hyperref}

\usepackage{amsmath}
\usepackage{amssymb}
\usepackage{mathtools}
\usepackage{amsthm}
\usepackage{kotex}

\usepackage[capitalize,noabbrev]{cleveref}

\theoremstyle{plain}

\theoremstyle{definition}

\theoremstyle{remark}

\usepackage[textsize=tiny]{todonotes}

\icmltitlerunning{Submission and Formatting Instructions for ICML 2023}

\begin{document}

\twocolumn[
\icmltitle{Synthetic Alone: Exploring the Dark Side of Synthetic Data for Grammatical Error Correction}



\icmlsetsymbol{equal}{*}

\begin{icmlauthorlist}
\icmlauthor{Chanjun Park}{yyy,yyyc,equal}
\icmlauthor{Seonmin Koo}{yyy,equal}
\icmlauthor{Seolhwa Lee}{yyyd,equal}
\icmlauthor{Jaehyung Seo}{yyy}
\icmlauthor{Sugyeong Eo}{yyy}
\icmlauthor{Hyeonseok Moon}{yyy}
\icmlauthor{Heuiseok Lim}{yyy}
\end{icmlauthorlist}

\icmlaffiliation{yyy}{Department of Computer Science and Engineering, Korea University, Seoul 02841, Korea}
\icmlaffiliation{yyyc}{Upstage, Gyeonggi-do, Korea}
\icmlaffiliation{yyyd}{Technical University of Darmstadt}


\icmlcorrespondingauthor{Heuiseok Lim}{limhseok@korea.ac.kr}

\icmlkeywords{Machine Learning, ICML}

\vskip 0.3in
]



\printAffiliationsAndNotice{\icmlEqualContribution} 

\begin{abstract}
Data-centric AI approach aims to enhance the model performance without modifying the model and has been shown to impact model performance positively. While recent attention has been given to data-centric AI based on synthetic data, due to its potential for performance improvement, data-centric AI has long been exclusively validated using real-world data and publicly available benchmark datasets. In respect of this, data-centric AI still highly depends on real-world data, and the verification of models using synthetic data has not yet been thoroughly carried out. Given the challenges above, we ask the question: \textit{``Does data quality control (noise injection and balanced data), a data-centric AI methodology acclaimed to have a positive impact, exhibit the same positive impact in models trained solely with synthetic data?''} To address this question, we conducted comparative analyses between models trained on synthetic and real-world data based on grammatical error correction (GEC) task. Our experimental results reveal that the data quality control method has a positive impact on models trained with real-world data, as previously reported in existing studies, while a {\em negative} impact is observed in models trained solely on synthetic data.
\end{abstract}

\section{Introduction}
Data-centric AI research has been actively conducted in natural language processing (NLP) to improve model performance without the need for significant cost and model modification. Several data-centric AI methods have been developed to achieve this goal, such as data management~\cite{choi2023dmops}, data filtering~\cite{koehn2020findings}, noise injection (perturbation)~\cite{sarp2021analysis,partovyan2018noise}, and data augmentation~\cite{shorten2019survey}. Among these methods, the use of synthetic data~\cite{nikolenko2019synthetic} has gained increasing interest with the development of the large language models (LLMs), such as GPT3~\cite{brown2020language}, ChatGPT\footnote{\url{https://chat.openai.com/}}, and  LaMDA~\cite{thoppilan2022lamda}, 
These LLMs have demonstrated the potential for generating high-quality synthetic data~\cite{chen2023places,wang2021towards} and the possibility of replacing the need for human-annotated data with synthetic data. 

\begin{table*}[htb!]
\centering
\scalebox{0.9}{
\begin{tabular}{@{}ll@{}}
\toprule
Correct sentence                      & \begin{tabular}[c]{@{}l@{}}바비큐 그릴도 이용할 수 있나요?\\ (Can I also use a barbecue grill?)\end{tabular}       \\ \midrule\midrule
Example of separation error           & \begin{tabular}[c]{@{}l@{}}ㅂㅏㅂㅣ큐 그릴도 이용할 수 있나요?\\ (Can I also use a b ar b e cue grill?)\end{tabular} \\\midrule
Example of vowel alteration error & \begin{tabular}[c]{@{}l@{}}바비큐 그릴됴 이용할 수 있나요?\\ (Can I alsa use a barbecue grill?)\end{tabular}      \\\midrule
Example of pronunciation error        & \begin{tabular}[c]{@{}l@{}}바비큐 그릴도 이용할 수 잇나요?\\ (Canai also use a barbecue grill?)\end{tabular}       \\\midrule
Example of punctuation attachment errors         & \begin{tabular}[c]{@{}l@{}}바비큐 그릴도. 이용할 수 있나요?\\ (Can I also. use a barbecue grill?)\end{tabular}     \\\midrule
Example of loanword error     & \begin{tabular}[c]{@{}l@{}}비비큐 그릴도 이용할 수 있나요?\\ (Can I also use a bebecue grill?)\end{tabular}       \\\midrule
Example of neologism error            & \begin{tabular}[c]{@{}l@{}}바비큐 그릴도 O.l용할 수 있나요?\\ (Can 1 also use a barbecue grill?)\end{tabular}    \\ \bottomrule
\end{tabular}}
\caption{Example of noise injection according to noise type.}
\label{tab:ex_noise}
\end{table*}

However, we raise questions on the validity of prior research in data-centric AI, which has been shown to impact model performance positively. Existing data-centric AI studies have basically been conducted based on human-annotated data or publicly open data. Still, validation of models using only synthetic data has not been sufficiently conducted~\cite{polyzotis2021can,mazumder2022dataperf}. 

In the data-centric AI research, studies have focused on efficient methods of generating synthetic data~\cite{park2021should} and human-like data~\cite{moon2022automatic}. However, there has been limited validation of model performance improvement through data quality control using fully synthetic data. This paper analyzes whether a model trained only on synthetic data rather than human-annotated data can still demonstrate a positive impact in a data-centric approach.

To do this, we employ the grammatical error correction (GEC) task as it is one of the closely related tasks to the real-world. We conduct experiments on the GEC task using two models: (1) a BackTranscription (BTS)~\cite{park2021bts}-based GEC model, which is a synthetic data generation method proposed in recent speech recognition post-processing, and (2) a GEC model for learning from real-world data~\cite{park2020comparison}. To analyze the impact of data quality control on performance, we apply methods such as noise injection~\cite{ivanovs2021perturbation} and balanced data~\cite{park2022mimicking,chen2021improving} to both models and compare their results.

Moreover, there have been studies on the effectiveness of synthetic data based on self-supervised learning~\cite{ng-etal-2020-ssmba, ruiter-etal-2021-integrating, gan2021self}. Also, through the scaling law that examines the performance of models based on the size of the dataset, it has been demonstrated to be highly effective to use data generated by models that increase the amount of data geometrically concerning the size of the dataset~\cite{jaiswal2020survey, raghunathan2021synthetic, kaplan2020scaling}. In respect of this, we aim to revisit the synthetic data research.

\section{Design for Revisiting the impact of Synthetic Data}
\label{ex_design}
We raise the following question to revisit the impact of synthetic data---\textit{``Does the data quality control manifest the same positive impact in models trained only on synthetic data?''}. To validate this question, we design the experiments from two different perspectives. We investigate the following questions through comparative analyses between the performance of the models trained on synthetic and real-world data. 



\begin{figure*}[h]
\begin{center}
\includegraphics[width=0.9 \linewidth]{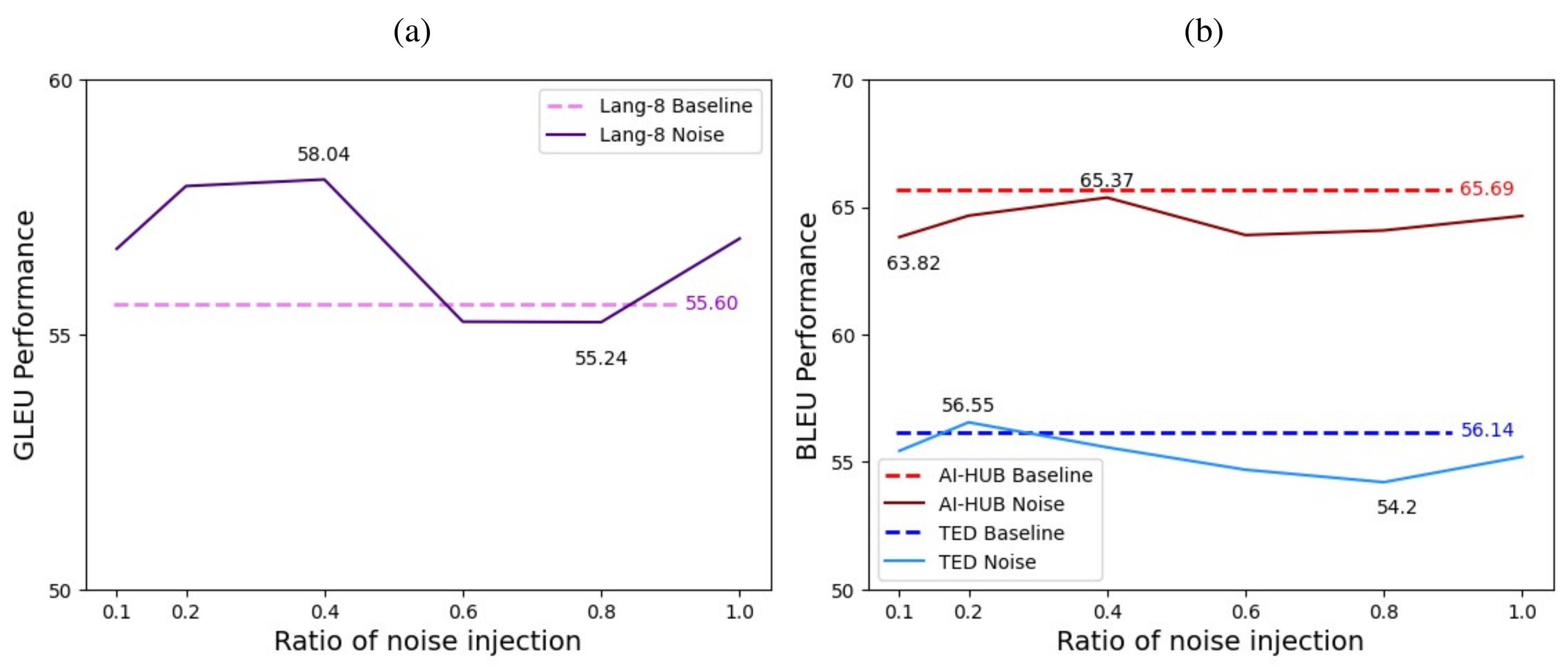}
\end{center}
\caption{Experimental results of noise injection. (a) is the result of inserting noise into real-world data. (b) is the result of inserting noise into synthetic data. Note that the x-axis indicates the strength of noise injection.}
\label{fig:main}
\end{figure*}

\begin{itemize}
    \item \textit{How does the strength of noise injection impact the model performance?}
\end{itemize}
To address this question, we propose employing the noise injection method, more specifically perturbation, which is representative of the data quality control method, to assess the performance between synthetic and real-world data.

For the perturbation of synthetic data BTS, six types of noise---i.e. separation, vowel alteration, pronunciation, punctuation attachment, loanword, and neologism errors---are applied to the source sentence. The separation error refers to the case where the consonant and vowels of a character are separated. The vowel alteration error is where the vowel of a character is replaced with a different vowel. The pronunciation error indicates a case where a character is altered by pronunciation. The punctuation attachment error refers to a case where punctuation is attached in an unnecessary position within a sentence. The loanword conversion error deals with cases where part of a character is converted into English. The neologism error refers to a case where the character is altered using grammar not included in the existing grammatical system. See detailed examples in Table~\ref{tab:ex_noise}.

Regarding perturbation for real-world data Lang-8, the same correct sentence pair is inserted into the source and target sentences. Due to the characteristics of the GEC task, the source sentence already contains errors, so it is considered noise to insert clean data into the source sentence, which should have noise that is opposed to the characteristics of the data.

Subsequently, the performance comparison between the baseline model trained on data without noise injection and the noised model trained on data with perturbation is conducted to examine the impact of noise injection on the model concerning synthetic and real-world data. Specifically, we conduct the experiment according to the strength of noise injection ranging from 0.1 to 1.0. Noise is inserted based on the ratio of noise set at the word level for each sentence. For example, if the noise ratio is 1.0, the noise will occur in all words within each sentence.

\begin{itemize}
    \item \textit{How does the ratio of noised and cleaned text batches impact the model performance?}
\end{itemize}

The aim is to obtain answers to the question by comparing the performance of synthetic data and real-world data by applying the balanced data method. The balanced data method is a method of training by intentionally giving appropriate ratios to noise and clean data. That is, when forming batches during training, data with different features is composed based on the pre-set ratios, and the training method is performed based on these ratios~\citet{park2022mimicking}. The performance experiment of the balance between clean and noisy data is conducted with five different ratios of synthetic and real-world data---5:5, 4:6, 3:7, 2:8, and 1:9. The comparison of the performance between the baseline model without any operations and the model trained with balanced data method is then carried out to analyze the impact of the ratio of noise and clean on the performance of the only synthetic and real-world data-based models.

\section{Experimental Settings}

\paragraph{Real-world \& Synthetic Data}
We use the Lang-8 dataset\footnote{\url{https://lang-8.com/}} as our real-world data, a fully human-annotated corpus. The data settings are consistent with those used by \citet{park2020comparison}.

We generate synthesized datasets fitting for the GEC task  from the above datasets (AI-HUB, TED) using BTS~\cite{park2021bts}. BTS combines text-to-speech (TTS) technology and speech-to-text (STT) technology to generate GEC task synthesized data for speech recognition post-processor. Although BTS cannot represent synthetic data, BTS is a simple and efficient methodology for generating synthetic data. Thus, we use it for experiments. As raw data for generating BTS-based synthetic data, we use AI-HUB~\cite{park2022empirical}, which are representative Korean data platforms, and TED Korean dataset\footnote{\url{https://www.ted.com/talks?language=ko}}, the same as existing BTS work. In addition, since the existing BTS research was also conducted in Korean, this experiment also performs based on Korean for a fair evaluation.

Table~\ref{tab:table_stat} shows the specific data statistics used in the experiment. We use 92,000 sentences from AI-HUB data and 119,883 sentences from TED's Korean Transcript data to generate BTS-based synthetic data followed by \citet{park2020comparison}'s method. These data are used as raw data for BTS, transformation into speech using TTS, and outputting the converted result as text using STT. 

\begin{table}[h]
{\small
  \centering
\begin{tabular}{@{}cccc@{}}
\toprule
 Dataset     & Train     & Test  & Type of data \\ \midrule
 Lang-8 & 1,075,513 & 631   & real-world          \\ \midrule
AI-HUB & 92,000    & 3,000 & synthetic(BTS based)         \\
TED   & 119,883   & 3,000 & synthetic(BTS based)           \\
\bottomrule
\end{tabular}
\caption{Statistics on the number of sentences according to real-world and synthetic data.}
\label{tab:table_stat}}
\end{table}

\paragraph{Implementation Details}
We train the models using the vanilla Transformer~\cite{vaswani2017attention} and set the same for hyperparameters. Fairseq~\cite{ott2019fairseq} is used for the implementation. For subword tokenization, we utilize SentencePiece ~\cite{kudo2018sentencepiece} and set the vocabulary size to 50,000. We evaluate the Lang-8-based real-world model and the BTS-based synthetic data model as GLEU~\cite{napoles2015ground} and BLEU~\cite{papineni2002bleu}, respectively. These are the same metrics as previous GEC and BTS papers.



\begin{figure*}[h]
\begin{center}
\includegraphics[width=0.9 \linewidth]{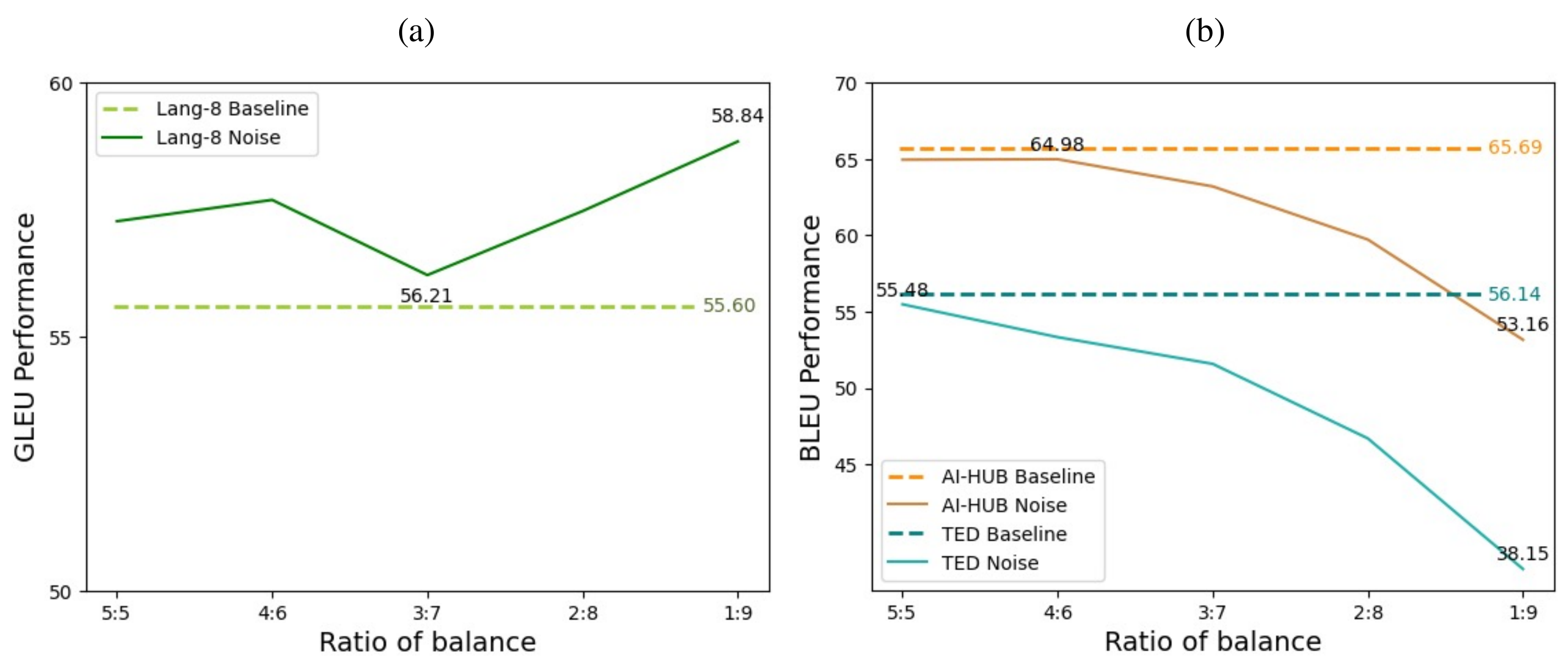}
\end{center}
\caption{Experimental results of balanced data. (a) is the balanced data result between noise and clean data in real-world data. (b) is the balanced data result between noise and clean data in synthetic data. Note that the x-axis indicates the ratio as (clean:noise).}
\label{fig:bal}
\end{figure*}
\section{Experimental Results}
\subsection{Results of Question 1: Noise Injection}
\label{main_res}
Figure~\ref{fig:main} shows the results of applying noise injection (perturbation) methods. Baseline for the real-world data (a) is 55.60. The performance tends to improve when injecting noise. Mainly, the 0.4 noise ratio result obtains a substantial gain of +2.44 to 58.04. Meanwhile, (b) indicates the experimental results of synthetic data on AI-HUB and TED datasets. 

We denote the probability of noise being injected into a token in a sentence as the noise injection ratio. The baselines of AI-HUB and TED report 65.69 and 56.14, respectively, and models learned with the data subject to noise injection show performance degradation in most cases. A marginal performance gain is recorded when the noise ratio of the TED dataset is 0.2; rather, the overall performance decreased except for this case. \textit{The results demonstrate that data quality control through noise injection, which is known to impact performance in many studies positively, has \textbf{a negative impact} when training a model using only synthetic data.}

These results starkly contrast with the experimental results using models trained entirely of real-world data, highlighting our conclusion of the \textit{\textbf{negative impact}} of the noise injection on the synthetic data. Namely, \textit{we recommend applying synthetic data after extensively verifying whether the data-centric AI methods are effective in a synthetic data-only setting.}

\subsection{Results of Question 2: Balanced Data}
The experimental results of applying balanced data methods are described in Figure~\ref{fig:bal}. As mentioned, we combine clean and noise data with a five ratio. The noise injection ratio on the noisy data is 1.0, the highest noise for Section~\ref{main_res}.

(a) is the experimental results for real-world data. Contrary to the synthetic data, all cases perform better than the baseline. The performance on the ratio of 1:9 is 58.84, showing a gain of 3.24 points from the baseline. As previously demonstrated, we confirm that leveraging data quality control techniques to real data positively impacts the model.

(b) is the result of the model training with synthetic data by intentionally giving ratios to clean and noise. As a result, the model learned with synthetic data performs worse than the baseline in all ratios. In particular, as the ratio of noise data increases, performance tends to deteriorate. Experimental results show \textit{\textbf{a negative effect overall}}, and \textit{we analyze that data quality control is less effective in model training in an environment consisting only of synthetic data.}

We conclude that data quality control positively affects real-world data-based models but does not always guarantee a positive effect in an environment consisting entirely of synthetic data. This implies that real-world and synthetic data have distinctly contrasting characteristics and should be treated differently. \textit{We argue that beyond effective synthetic data generation, which is the focus of the data-centric AI, data quality control methods should be inspected to ensure that models leveraging synthetic data produce sufficient performance.}

\section{Conclusion and Future Work}
In this work, we address the research question of \textit{``Is the data quality control, a data-centric AI methodology known to have a positive effect, still observed to have a positive impact when the models are trained only using synthetic data?''} by performing experiments and evaluating the results. Our experiments reveal that while the conventional data-centric approach positively impacted real-world data, models trained solely on synthetic data showed a negative impact. This demonstrates that data-centric methodologies do not necessarily guarantee positive effects, dependent on the characteristics of the data. Based on the results, it was found that sufficient evaluation of data-centric methods in synthetic data environments is needed. The experimental results cannot be generalized since the experiment was limited to the GEC task and not all data-centric AI methodologies were tested. However, a clearly defined research question was addressed, and insightful results were obtained through a structured comparison. In future work, we aim to further analyze the characteristics of synthetic data environments by implementing data-centric approaches other than noise injection and balanced data.


\section*{Acknowledgements}
This research was supported by the MSIT(Ministry of Science and ICT), Korea, under the ITRC(Information Technology Research Center) support program(IITP-2023-2018-0-01405) supervised by the IITP(Institute for Information \& Communications Technology Planning \& Evaluation), and This work was supported by the National Research Foundation of Korea(NRF) grant funded by the Korea government(MSIT)(No. 2022R1A5A7026673).

\nocite{langley00}

\bibliography{example_paper}

\begin{thebibliography}{31}
\providecommand{\natexlab}[1]{#1}
\providecommand{\url}[1]{\texttt{#1}}
\expandafter\ifx\csname urlstyle\endcsname\relax
  \providecommand{\doi}[1]{doi: #1}\else
  \providecommand{\doi}{doi: \begingroup \urlstyle{rm}\Url}\fi

\bibitem[Brown et~al.(2020)Brown, Mann, Ryder, Subbiah, Kaplan, Dhariwal,
  Neelakantan, Shyam, Sastry, Askell, et~al.]{brown2020language}
Brown, T., Mann, B., Ryder, N., Subbiah, M., Kaplan, J.~D., Dhariwal, P.,
  Neelakantan, A., Shyam, P., Sastry, G., Askell, A., et~al.
\newblock Language models are few-shot learners.
\newblock \emph{Advances in neural information processing systems},
  33:\penalty0 1877--1901, 2020.

\bibitem[Chen et~al.(2021)Chen, Wan, and Dou]{chen2021improving}
Chen, L., Wan, S., and Dou, L.
\newblock Improving diagnostic performance of high-voltage circuit breakers on
  imbalanced data using an oversampling method.
\newblock \emph{IEEE Transactions on Power Delivery}, 37\penalty0 (4):\penalty0
  2704--2716, 2021.

\bibitem[Chen et~al.(2023)Chen, Papangelis, Tao, Kim, Rosenbaum, Liu, Yu, and
  Hakkani-Tur]{chen2023places}
Chen, M., Papangelis, A., Tao, C., Kim, S., Rosenbaum, A., Liu, Y., Yu, Z., and
  Hakkani-Tur, D.
\newblock Places: Prompting language models for social conversation synthesis.
\newblock \emph{Proceedings of the 17th Conference of the European Chapter of
  the Association for Computational Linguistics}, 2023.

\bibitem[Choi \& Park(2023)Choi and Park]{choi2023dmops}
Choi, E. and Park, C.
\newblock Dmops: Data management operation and recipes.
\newblock \emph{arXiv preprint arXiv:2301.01228}, 2023.

\bibitem[Gan et~al.(2021)Gan, Xu, and Zan]{gan2021self}
Gan, Z., Xu, H., and Zan, H.
\newblock Self-supervised curriculum learning for spelling error correction.
\newblock In \emph{Proceedings of the 2021 Conference on Empirical Methods in
  Natural Language Processing}, pp.\  3487--3494, 2021.

\bibitem[Ivanovs et~al.(2021)Ivanovs, Kadikis, and
  Ozols]{ivanovs2021perturbation}
Ivanovs, M., Kadikis, R., and Ozols, K.
\newblock Perturbation-based methods for explaining deep neural networks: A
  survey.
\newblock \emph{Pattern Recognition Letters}, 150:\penalty0 228--234, 2021.

\bibitem[Jaiswal et~al.(2020)Jaiswal, Babu, Zadeh, Banerjee, and
  Makedon]{jaiswal2020survey}
Jaiswal, A., Babu, A.~R., Zadeh, M.~Z., Banerjee, D., and Makedon, F.
\newblock A survey on contrastive self-supervised learning.
\newblock \emph{Technologies}, 9\penalty0 (1):\penalty0 2, 2020.

\bibitem[Kaplan et~al.(2020)Kaplan, McCandlish, Henighan, Brown, Chess, Child,
  Gray, Radford, Wu, and Amodei]{kaplan2020scaling}
Kaplan, J., McCandlish, S., Henighan, T., Brown, T.~B., Chess, B., Child, R.,
  Gray, S., Radford, A., Wu, J., and Amodei, D.
\newblock Scaling laws for neural language models.
\newblock \emph{arXiv preprint arXiv:2001.08361}, 2020.

\bibitem[Koehn et~al.(2020)Koehn, Chaudhary, El-Kishky, Goyal, Chen, and
  Guzm{\'a}n]{koehn2020findings}
Koehn, P., Chaudhary, V., El-Kishky, A., Goyal, N., Chen, P.-J., and
  Guzm{\'a}n, F.
\newblock Findings of the wmt 2020 shared task on parallel corpus filtering and
  alignment.
\newblock In \emph{Proceedings of the Fifth Conference on Machine Translation},
  pp.\  726--742, 2020.

\bibitem[Kudo \& Richardson(2018)Kudo and Richardson]{kudo2018sentencepiece}
Kudo, T. and Richardson, J.
\newblock Sentencepiece: A simple and language independent subword tokenizer
  and detokenizer for neural text processing.
\newblock \emph{arXiv preprint arXiv:1808.06226}, 2018.

\bibitem[Mazumder et~al.(2022)Mazumder, Banbury, Yao, Karla{\v{s}}, Rojas,
  Diamos, Diamos, He, Kiela, Jurado, et~al.]{mazumder2022dataperf}
Mazumder, M., Banbury, C., Yao, X., Karla{\v{s}}, B., Rojas, W.~G., Diamos, S.,
  Diamos, G., He, L., Kiela, D., Jurado, D., et~al.
\newblock Dataperf: Benchmarks for data-centric ai development.
\newblock \emph{arXiv preprint arXiv:2207.10062}, 2022.

\bibitem[Moon et~al.(2022)Moon, Park, Seo, Eo, and Lim]{moon2022automatic}
Moon, H., Park, C., Seo, J., Eo, S., and Lim, H.
\newblock An automatic post editing with efficient and simple data generation
  method.
\newblock \emph{IEEE Access}, 10:\penalty0 21032--21040, 2022.

\bibitem[Napoles et~al.(2015)Napoles, Sakaguchi, Post, and
  Tetreault]{napoles2015ground}
Napoles, C., Sakaguchi, K., Post, M., and Tetreault, J.
\newblock Ground truth for grammatical error correction metrics.
\newblock In \emph{Proceedings of the 53rd Annual Meeting of the Association
  for Computational Linguistics and the 7th International Joint Conference on
  Natural Language Processing (Volume 2: Short Papers)}, pp.\  588--593, 2015.

\bibitem[Ng et~al.(2020)Ng, Cho, and Ghassemi]{ng-etal-2020-ssmba}
Ng, N., Cho, K., and Ghassemi, M.
\newblock {SSMBA}: Self-supervised manifold based data augmentation for
  improving out-of-domain robustness.
\newblock In \emph{Proceedings of the 2020 Conference on Empirical Methods in
  Natural Language Processing (EMNLP)}, pp.\  1268--1283, Online, November
  2020. Association for Computational Linguistics.
\newblock \doi{10.18653/v1/2020.emnlp-main.97}.
\newblock URL \url{https://aclanthology.org/2020.emnlp-main.97}.

\bibitem[Nikolenko(2019)]{nikolenko2019synthetic}
Nikolenko, S.~I.
\newblock Synthetic data for deep learning.
\newblock \emph{arXiv preprint arXiv:1909.11512}, 2019.

\bibitem[Ott et~al.(2019)Ott, Edunov, Baevski, Fan, Gross, Ng, Grangier, and
  Auli]{ott2019fairseq}
Ott, M., Edunov, S., Baevski, A., Fan, A., Gross, S., Ng, N., Grangier, D., and
  Auli, M.
\newblock fairseq: A fast, extensible toolkit for sequence modeling.
\newblock \emph{arXiv preprint arXiv:1904.01038}, 2019.

\bibitem[Papineni et~al.(2002)Papineni, Roukos, Ward, and
  Zhu]{papineni2002bleu}
Papineni, K., Roukos, S., Ward, T., and Zhu, W.-J.
\newblock Bleu: a method for automatic evaluation of machine translation.
\newblock In \emph{Proceedings of the 40th annual meeting of the Association
  for Computational Linguistics}, pp.\  311--318, 2002.

\bibitem[Park et~al.(2020)Park, Yang, Lee, and Lim]{park2020comparison}
Park, C., Yang, Y., Lee, C., and Lim, H.
\newblock Comparison of the evaluation metrics for neural grammatical error
  correction with overcorrection.
\newblock \emph{IEEE Access}, 8:\penalty0 106264--106272, 2020.

\bibitem[Park et~al.(2021{\natexlab{a}})Park, Lee, Moon, Eo, Seo, and
  Lim]{park2021should}
Park, C., Lee, S., Moon, H., Eo, S., Seo, J., and Lim, H.
\newblock How should human translation coexist with nmt? efficient tool for
  building high quality parallel corpus.
\newblock \emph{arXiv preprint arXiv:2111.00191}, 2021{\natexlab{a}}.

\bibitem[Park et~al.(2021{\natexlab{b}})Park, Seo, Lee, Lee, Moon, Eo, and
  Lim]{park2021bts}
Park, C., Seo, J., Lee, S., Lee, C., Moon, H., Eo, S., and Lim, H.-S.
\newblock Bts: Back transcription for speech-to-text post-processor using
  text-to-speech-to-text.
\newblock In \emph{Proceedings of the 8th Workshop on Asian Translation
  (WAT2021)}, pp.\  106--116, 2021{\natexlab{b}}.

\bibitem[Park et~al.(2022{\natexlab{a}})Park, Go, Eo, Moon, Lee, and
  Lim]{park2022mimicking}
Park, C., Go, W.-Y., Eo, S., Moon, H., Lee, S., and Lim, H.
\newblock Mimicking infants’ bilingual language acquisition for domain
  specialized neural machine translation.
\newblock \emph{IEEE Access}, 10:\penalty0 38684--38693, 2022{\natexlab{a}}.

\bibitem[Park et~al.(2022{\natexlab{b}})Park, Shim, Eo, Lee, Seo, Moon, and
  Lim]{park2022empirical}
Park, C., Shim, M., Eo, S., Lee, S., Seo, J., Moon, H., and Lim, H.
\newblock Empirical analysis of parallel corpora and in-depth analysis using
  liwc.
\newblock \emph{Applied Sciences}, 12\penalty0 (11):\penalty0 5545,
  2022{\natexlab{b}}.

\bibitem[Partovyan et~al.(2018)Partovyan, Nourani, and
  ALAMI]{partovyan2018noise}
Partovyan, A., Nourani, V., and ALAMI, M.~T.
\newblock Noise injection--denoising techniques to improve artificial
  intelligence-based rainfall--runoff modeling.
\newblock \emph{Water Resources Engineering}, 11\penalty0 (36):\penalty0
  81--94, 2018.

\bibitem[Polyzotis \& Zaharia(2021)Polyzotis and Zaharia]{polyzotis2021can}
Polyzotis, N. and Zaharia, M.
\newblock What can data-centric ai learn from data and ml engineering?
\newblock \emph{arXiv preprint arXiv:2112.06439}, 2021.

\bibitem[Raghunathan(2021)]{raghunathan2021synthetic}
Raghunathan, T.~E.
\newblock Synthetic data.
\newblock \emph{Annual review of statistics and its application}, 8:\penalty0
  129--140, 2021.

\bibitem[Ruiter et~al.(2021)Ruiter, Klakow, van Genabith, and
  Espa{\~n}a-Bonet]{ruiter-etal-2021-integrating}
Ruiter, D., Klakow, D., van Genabith, J., and Espa{\~n}a-Bonet, C.
\newblock Integrating unsupervised data generation into self-supervised neural
  machine translation for low-resource languages.
\newblock In \emph{Proceedings of Machine Translation Summit XVIII: Research
  Track}, pp.\  76--91, Virtual, August 2021. Association for Machine
  Translation in the Americas.
\newblock URL \url{https://aclanthology.org/2021.mtsummit-research.7}.

\bibitem[Sarp et~al.(2021)Sarp, Kuzlu, Cali, Elma, and Guler]{sarp2021analysis}
Sarp, S., Kuzlu, M., Cali, U., Elma, O., and Guler, O.
\newblock Analysis of false data injection impact on ai based solar
  photovoltaic power generation forecasting.
\newblock \emph{arXiv preprint arXiv:2110.09948}, 2021.

\bibitem[Shorten \& Khoshgoftaar(2019)Shorten and
  Khoshgoftaar]{shorten2019survey}
Shorten, C. and Khoshgoftaar, T.~M.
\newblock A survey on image data augmentation for deep learning.
\newblock \emph{Journal of big data}, 6\penalty0 (1):\penalty0 1--48, 2019.

\bibitem[Thoppilan et~al.(2022)Thoppilan, De~Freitas, Hall, Shazeer,
  Kulshreshtha, Cheng, Jin, Bos, Baker, Du, et~al.]{thoppilan2022lamda}
Thoppilan, R., De~Freitas, D., Hall, J., Shazeer, N., Kulshreshtha, A., Cheng,
  H.-T., Jin, A., Bos, T., Baker, L., Du, Y., et~al.
\newblock Lamda: Language models for dialog applications.
\newblock \emph{arXiv preprint arXiv:2201.08239}, 2022.

\bibitem[Vaswani et~al.(2017)Vaswani, Shazeer, Parmar, Uszkoreit, Jones, Gomez,
  Kaiser, and Polosukhin]{vaswani2017attention}
Vaswani, A., Shazeer, N., Parmar, N., Uszkoreit, J., Jones, L., Gomez, A.~N.,
  Kaiser, {\L}., and Polosukhin, I.
\newblock Attention is all you need.
\newblock \emph{Advances in neural information processing systems}, 30, 2017.

\bibitem[Wang et~al.(2021)Wang, Yu, Firat, and Cao]{wang2021towards}
Wang, Z., Yu, A.~W., Firat, O., and Cao, Y.
\newblock Towards zero-label language learning.
\newblock \emph{arXiv preprint arXiv:2109.09193}, 2021.

\end{thebibliography}
\bibliographystyle{icml2023}

\end{document}